# Accelerated Reconstruction of Perfusion-Weighted MRI Enforcing Jointly Local and Nonlocal Spatio-temporal Constraints

Cagdas Ulas*, Christine Preibisch, Jonathan Sperl, Thomas Pyka, Jayashree Kalpathy-Cramer and Bjoern Menze

*Abstract*—Perfusion-weighted magnetic resonance imaging (MRI) is an imaging technique that allows one to measure tissue perfusion in an organ of interest through the injection of an intravascular paramagnetic contrast agent (CA). Due to a preference for high temporal and spatial resolution in many applications, this modality could significantly benefit from accelerated data acquisitions. In this paper, we specifically address the problem of reconstructing perfusion MR image series from a subset of k-space data. Our proposed approach is motivated by the observation that temporal variations (dynamics) in perfusion imaging often exhibit correlation across different spatial scales. Hence, we propose a model that jointly penalizes the voxel-wise deviations in temporal gradient images obtained based on a baseline, and the patch-wise dissimilarities between the spatio-temporal neighborhoods of entire image sequence. We validate our method on dynamic susceptibility contrast (DSC)-MRI and dynamic contrast-enhanced (DCE)-MRI brain perfusion datasets acquired from 10 tumor patients in total. We provide extensive analysis of reconstruction performance and perfusion parameter estimation in comparison to state-of-the-art reconstruction methods. Experimental results on clinical datasets demonstrate that our reconstruction model can potentially achieve up to 8-fold acceleration by enabling accurate estimation of perfusion parameters while preserving spatial image details and reconstructing the complete perfusion time-intensity curves (TICs).

## I. INTRODUCTION

STUDYING blood flow and blood flow patterns is a major field in clinical radiology and diagnostics. Perfusion-weighted magnetic resonance imaging (MRI) provides a mean for assessing tissue perfusion and vascular permeability *in vivo* through examination of the spatio-temporal changes of signal intensities following the injection of an exogenous paramagnetic contrast agent (CA) [1]. These techniques have become valuable clinical tools since they play a crucial role, for instance, in the diagnosis of stroke, the determination of tissue(s) at risk of infarction, and the prediction of prognosis after treatments of patients with stroke and tumors [2]. Two of the most common methods used in perfusion-weighted imaging (PWI) are dynamic susceptibility contrast MRI (DSC-MRI) and dynamic contrast enhanced MRI (DCE-MRI).

C. Ulas* and B. Menze are with the Department of Computer Science at the Technical University of Munich, Germany (e-mail: cagdas.ulas@tum.de).

C. Preibisch and T. Pyka are with the Department of Neuroradiology at the Technical University of Munich, Germany.

C. Ulas* and J. Sperl are with GE Global Research, Garching, Germany.

J. Kalpathy-Cramer is with the Department of Radiology at Harvard Medical School and the Department of Neuroscience at Massachusetts General Hospital, Boston, USA.

Both techniques require intravenous bolus administration of gadolinium, followed by the acquisition of successive images as the contrast bolus enters and subsequently leaves the organ of interest. DSC-MRI relies on dynamic alterations of the $T_2^*$ transverse relaxation times of tissues and it is employed to assess the hemodynamic status of tissues [3]. DCE-MRI on the other hand relies on changes of the $T_1$ longitudinal relaxation times of the tissues. DCE-MRI is widely used to interrogate the vascular characteristics of tumors in clinical settings [2].

Vast majority of clinical research on PWI have considered the problem of estimating accurate voxel-wise perfusion parameters which are generally obtained by fitting a tracer kinetic model to the observed time-intensity curves (TICs) [4]. An illustration displaying the major steps of kinetic parameter estimation in DSC-MRI is provided in supplementary material.

One of the major obstacles in the clinical use of PWI is the immense need of high temporal resolution to capture the rapid contrast changes of CA uptake for precise perfusion quantification [1]. Furthermore, the short scanning time available for each frame often leads to limited spatial resolution to detect small image features and accurate tumor boundaries, and low signal-to-noise ratio (SNR) to enable precise fitting of kinetic model parameters. Considering such severe constraints, PWI can benefit from subsampled acquisitions [5]. However, sub-Nyquist sampling often results in aliasing artifacts in image space and in the context of PWI, reconstruction of complete temporal signal with its peak and high dynamics (observable in blood vessels) constitutes even a more challenging problem.

Recently, various reconstruction approaches have been proposed in related dynamic imaging applications, based on, such as piece-wise smoothness in the spatial domain [6], [7], high correlation and sparsity in the temporal domain [7], [8], [9], [10], sparse representations of local image regions via learned dictionaries [10], [11] and low-rank property of MR sequences in the full spatio-temporal space [8], [12], [6]. However, there are only a few works dedicated directly to reconstruction problem in PWI, considering the constraint of the image frames based on a baseline (pre-contrast) image [13], the penalization of time curves with high temporal gradients [14], and the minimization of temporal finite-differences enforced together with multiple constraints on spatial domain [7], [15]. The main limitation of these methods is that they consider the temporal variations only in single voxel level and neglect the similarities and variations between the voxels located in a spatially close neighborhood. For this reason, their performance is very sensitive to rapid signal changes occurring over the duration of image acquisition as encountered in PWI. These methods often produce blurry image regions and oversmooth reconstruction of TICs, and ultimately result in underestimated peak value of the perfusion signal, which substantially deteriorates the accuracy of estimated perfusion parameters. To this end, an optimal choice of the reconstruction model is essential for PWI targeting to recover the complete temporal pattern of the perfusion TICs while preserving the spatial quality of image series. We believe that a reconstruction model satisfying such conditions can be used to accelerate the acquisition of PWI and practically allow to acquire more samples in time domain and improve the volume coverage of the organ of interest.

In perfusion imaging, we observe rapid signal (contrast) variations over time due to the attenuation related to the passage of the contrast agent. Depending on the level of perfusion incurred by the tracer, these variations mainly differ in tissues, blood vessels, and tumor regions, etc., appearing as small local areas inside the organ to be imaged. Motivated by this observation, we propose a new reconstruction model specifically for PWI. Our model primarily integrates two fundamentally different data-driven constraints: (i) a voxel-wise local sparsity constraint on the temporal gradient images with respect to a baseline, limiting the overall dynamic of the perfusion time series, and (ii) a patch-wise similarity constraint on the spatio-temporal neighborhoods of the entire perfusion image series, providing smooth spatial regions with better alignment to temporal variations in small local areas represented with patches. We formulate the main optimization problem in a joint formal framework and introduce a new proximal splitting strategy [16] that benefits from the weighted-average of proximals, and can efficiently solve the joint minimization problem with fast convergence. The proposed method is validated on DSC and DCE-MRI perfusion datasets collected from brain tumor patients and compared with existing reconstruction methods. Extensive experiments demonstrate the efficiency of our method in terms of reconstruction performance and estimation of perfusion parameters from accelerated acquisitions. To the best of our knowledge, this is the first work to exploit the spatial and temporal variations jointly at different scales for the purpose of reconstruction of PWI, successfully applied on both DSC and DCE-MRI time series.

Preliminary results of this work presented at a conference [17] are herein extended by additional validation on DCE-MRI datasets collected from a clinical cohort of glioma patients, assessment of our method with estimated hemodynamic and pharmacokinetic parameters underlying perfusion, and experimental analysis of convergence properties of the proposed algorithm. The key contributions of this paper can be summarized as follows:

- We present a robust reconstruction method for PWI dynamic series. Our proposed model exploits the spatiotemporal variations jointly at single voxel and patch level.
- The proposed reconstruction model can enable accurate quantification of clinically useful perfusion parameters while attaining up to 8-fold acceleration through the use of only a subset of k-space measurements.
- We introduce a formal iterative algorithm to solve the minimization of the sum of convex regularizers based on

proximal-splitting applied to a reconstruction problem.

The remainder of this paper is organized as follows. Section II provides the detailed description of the proposed reconstruction model along with its formulation and the algorithm to solve the reconstruction problem. In Section III, we briefly describe the acquisition parameters of our clinical perfusion datasets and employed tracer kinetic models for data analysis. Section IV presents the experimental setup and the results of conducted experiments. After a general discussion we provide the concluding remarks in Section V.

## II. RECONSTRUCTION MODEL

Our proposed reconstruction model jointly imposes two spatio-temporal constraints both on a voxel-wise (local) and patch-wise (nonlocal) level as illustrated in Fig. 1. In the following sections we provide the intuition behind using specifically these constraints and describe how to mathematically formulate the joint regularization problem along with the algorithm that efficiently solves the optimization.

### A. Formulation

We remark that throughout the paper we describe our method on 2D + t data only for simplicity of the presentation. However, a generalization to 3D + t volumes is also straightforward. We assume that $X \in \mathbb{R}^{N_x \times N_y \times T}$ is a 2D perfusion MR image series (sequence) represented as spatio-temporal 3D data involving a total number of $N = N_x \times N_y \times T$ voxels. Let $x_t \in \mathbb{R}^{N_x \times N_y}$ denote a perfusion MR frame at time $t$, $y_t$ is the acquired undersampled k-space measurement of $x_t$, and $\mathbb{T} = \{1, 2, ..., T\}$ is the set of frame number indices in the sequence. The main objective is to reconstruct all $x_t$'s from the acquired k-space measurements $y_t$'s. The physical model between $x_t$ and $y_t$ can be formulated as,

$$y_t = A_t(F x_t + \eta),$$ (1)

where $A_t$ represent the sampling matrix to acquire only a subset of k-space samples, $F$ is the 2D Fourier Transform operator and $\eta$ is additive Gaussian noise in k-space. We denote the partial 2D Fourier operator for frame $t$ as $\mathcal{F}_t = A_t F$, and stack the $\mathcal{F}_t$'s for all frames of the sequence as $\mathcal{F}_u = \text{diag}\{\mathcal{F}_1, \mathcal{F}_2, .., \mathcal{F}_T\}$. The investigated problem in (1) is an ill-posed inverse problem [6], [10]. Regularization is often required to find a unique and stable solution to such problems.

We pose the joint regularization for the reconstruction of perfusion image series as the following optimization problem,

$$\hat{X} = \arg\min_{X} \ \frac{1}{2} \|\mathcal{F}_u X - Y\|_2^2 + \lambda_1 \mathcal{R}_L(X) + \lambda_2 \mathcal{R}_{NL}(X),$$ (2)

where $X$ denotes the entire perfusion image series and $Y$ represents the corresponding k-space measurements. The first term in (2) ensures data consistency, $\mathcal{R}_L$ and $\mathcal{R}_{NL}$ are two regularization terms imposed on reconstruction, $\lambda_1$ and $\lambda_2$ are the tuning parameters for these penalty terms.

**Local ($\mathcal{R}_L$) regularizer:** This regularizer penalizes the sum of voxel-wise intensity differences in temporal gradients calculated based on a reference for every frame $x_t$, defined as,

$$\mathcal{R}_L(X) = \sum_{t \in \mathbb{T}} \sum_{n=1}^{N_x \times N_y} \sqrt{(\nabla_x (x_t - \bar{x})_n)^2 + (\nabla_y (x_t - \bar{x})_n)^2},$$

where $\bar{x}$ is a reference (baseline) image, $\nabla_x$ and $\nabla_y$ represent finite-difference operators along $x$ and $y$ dimensions, respectively. This penalty is termed as dynamic total variation (DTV) and was orignally proposed for online reconstruction [9]. Here we employ DTV in offline manner, where all the frames are available. This term is better adjusted to the variations in time since it explicitly enforces temporal coherence by minimizing the difference with respect to a reference for every frame. Presumably, if there are deviations from the baseline intensities, they should be spatially homogeneous, i.e., a block of neighboring voxels should exhibit the same amount of deviation. Intuitively, this regularizer serves as a penalty on the large deviations from a baseline perfusion signal, enabling to smooth extreme local image regions.

***Nonlocal ($\mathcal{R}_{NL}$) regularizer:*** This regularizer penalizes the weighted sum of $\ell_2$ norm distances between spatio-temporal neighborhoods (patches) of the image series and we use it as a fully 3D nonlocal scheme. The term is specified by [18],

$$\mathcal{R}_{NL}(X) = \sum_{\mathbf{p} \in \Omega} \sum_{\mathbf{q} \in \mathcal{N}_{\mathbf{p}}} \varphi(\mathbf{p}, \mathbf{q}) \| P_{\mathbf{p}}(X) - P_{\mathbf{q}}(X) \|_2^2,$$

where $\mathbf{p} = (p_x, p_y, p_t)$ and $\mathbf{q} = (q_x, q_y, q_t)$ are two voxels, and the voxel of interest is $\mathbf{p} \in \Omega$, where $\Omega = [0, N_x] \times [0, N_y] \times [0, T]$. The term $P_{\mathbf{p}}(X)$ denotes a spatio-temporal 3D patch of the image sequence X, centered at voxel $\mathbf{p}$. We represent $\mathcal{N}_{\mathbf{p}}$ as a 3D search window around voxel $\mathbf{p}$. We simply denote $N_p$ and $N_w$ as the size of a patch and search window, respectively. The size of the patch must be smaller than the size of the search window. The weights $\varphi(\mathbf{p}, \mathbf{q})$ are calculated based on Euclidean distance between the patches,

$$\varphi(\mathbf{p}, \mathbf{q}) = \exp\left(-\|P_{\mathbf{p}}(X) - P_{\mathbf{q}}(X)\|_2^2 / h^2\right), \quad (3)$$

where $h$ is a parameter controlling the decay of the exponential function. The exponential weighting favors the similar patches in terms of Euclidean distance by assigning higher weight to their center voxels. Intuitively, this regularizer can exploit the similarities between patch pairs in adjacent frames and enforce smooth solutions in the spatio-temporal neighborhoods of the MR sequence even in the presence of significant inter-frame signal changes and high noise introduced during acquisition.

### B. Algorithm

The algorithm solving the primal problem (2) is mainly based on a proximal-splitting framework. For the better understanding of our algorithm, we first start with the definition of a proximal map.

***Proximal map:*** Given a continuous convex function $g(x)$ and a scalar $\rho > 0$, the proximal operator associated to convex function $g$ can be defined as [19]

$$prox_\rho(g)(z) := \underset{x \in \mathcal{H}}{\arg\min} \ \frac{1}{2\rho} \|x - z\|_2^2 + g(x). \quad (4)$$

Concretely, this operator serves as an individual minimizer for the convex function $g$, which approximates a value close to a reference point $z$ [20].

The reconstruction problem in (2) can be reformulated as the following denoising problem,

$$\hat{X} = \underset{X}{\arg\min} \ \frac{1}{2} \|X - Z\|_2^2 + \rho\lambda_1 \mathcal{R}_L(X) + \rho\lambda_2 \mathcal{R}_{NL}(X), \quad (5)$$

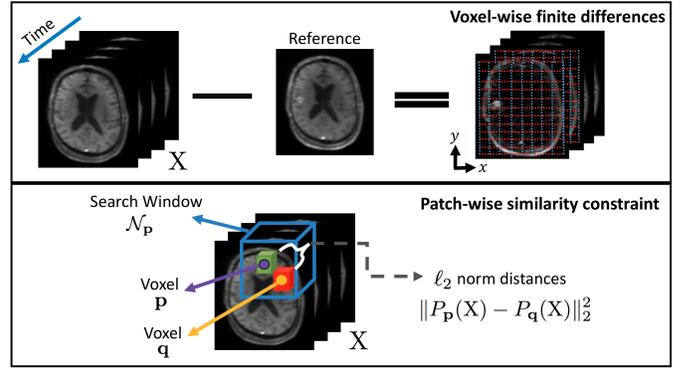

Fig. 1. Illustration of the two constraints enforced in our reconstruction model.

where $Z = \mathcal{F}_u^H Y$ and $H$ is the conjugate transpose. Assuming that each of the regularization terms in the cost function (2) is convex, the above denoising problem can be represented as the proximal map of the sum of two regularization terms as described in [19],

$$\hat{X} = prox_\rho(\lambda_1 \mathcal{R}_L + \lambda_2 \mathcal{R}_{NL})(Z). \quad (6)$$

The problem (5) admits to a unique solution as provided in (6). Proximal-splitting methods can allow tractable solutions for the proximity operator of the sum of two convex functions. These methods are first-order iterative algorithms that solve relatively large-scale optimization problems with several nonsmooth penalties. They operate by splitting the main objective function into individual subproblems which can be easily evaluated via proximal operators [20].

To solve our main problem in (5), we therefore split the objective function into two individual subproblems that we term $\mathcal{R}_L$-subproblem and $\mathcal{R}_{NL}$-subproblem.

***$\mathcal{R}_L$-subproblem:*** The proximal map for this subproblem can be defined as,

$$prox_\rho(\lambda_1 \mathcal{R}_L)(Z) = \underset{X}{\arg\min} \ \frac{1}{2\rho} \|X - Z\|_2^2 + \lambda_1 \mathcal{R}_L(X).$$

To efficiently solve this subproblem, we first reformulate it by introducing new variables $d_t = x_t - \bar{x}$, $Z_t = \mathcal{F}_t^H y_t$ and $d_g^t = Z_t - \bar{x}$, then the problem can be turned into

$$\hat{d} = \underset{d}{\arg\min} \ \sum_{t \in \mathbb{T}} \left( \frac{1}{2\rho} \|d_t - d_g^t\|_2^2 + \lambda_1 \|d_t\|_{TV} \right), \quad (7)$$

where $d = \{d_1, ..., d_T\}$ and $\|d_t\|_{TV} = \|[Q_1 d_t, Q_2 d_t]\|_{2,1}$, where $Q_1$ and $Q_2$ are two $N_x N_y \times N_x N_y$ first order finite difference matrices in vertical and horizontal directions, $\ell_{2,1}$ norm is the sum of the $\ell_2$ norm of each row of given matrix.

When a reference image $\bar{x}$ is given, the cost function in (7) can be minimized individually for every frame $x_t$ [9]. This guarantees that the sum of the costs is also minimized. The minimization can be efficiently solved using the fast iteratively reweighted least squares (FIRLS) algorithm [21] based on preconditioned conjugate gradient method. This algorithm provides fast convergence and low computational cost by adopting a preconditioner approximated using diagonally dominant structure of the symmetric matrix $\mathcal{F}_t^H \mathcal{F}_t$. Once the problem (7) is solved, the final solution of $\mathcal{R}_L$-subproblem is simply obtained by,

$$\hat{X}_{\mathcal{R}_L} = \left[ \hat{d}_1 + \bar{x}, \hat{d}_2 + \bar{x}, ...., \hat{d}_T + \bar{x} \right]. \quad (8)$$

---

**Algorithm 1:** Reconstruction algorithm

---

**Input:** Acquired k-space data $Y$, $\mathcal{F}_u$, $\lambda_1$, $\lambda_2$

**Initialize:** $z_1^0 = z_2^0 = \mathcal{F}_u^H Y$, $w_1$, $w_2$, $X^0 = \sum_{i=1}^{2} w_i z_i^0$, $\alpha_0$,
$\gamma = 1$, $k = 0$

**while** *stopping criteria not met* **do**

$\quad X_g = X^k - \gamma \mathcal{F}_u^H (\mathcal{F}_u X^k - Y)$

$\quad z_1^{k+1} = z_1^k + \alpha_k (prox_{\frac{\gamma}{w_1}} (2\lambda_1 \mathcal{R}_L)(X^k + X_g - z_1^k) - X^k)$

$\quad z_2^{k+1} = z_2^k + \alpha_k (prox_{\frac{\gamma}{w_2}} (2\lambda_2 \mathcal{R}_{NL})(X^k + X_g - z_2^k) - X^k)$

$\quad X^{k+1} = w_1 z_1^{k+1} + w_2 z_2^{k+1}$

$\quad \alpha_{k+1} = 1 + (2\alpha_k - 1)/(1 + \sqrt{1 + 4(\alpha_k)^2})$

$\quad k \leftarrow k + 1$

**end**

**Output:** Reconstructed image sequence $X$

---

$\mathcal{R}_{NL}$-*subproblem:* The proximal map for this subproblem can be specified by,

$$prox_\rho(\lambda_2 \mathcal{R}_{NL})(Z) = \arg\min_X \frac{1}{2\rho} \|X - Z\|_2^2 + \lambda_2 \mathcal{R}_{NL}(X).$$

The nonlocal penalty function in this problem is nonconvex. However, it has been shown in [22] that the problem has a convex nature when the nonlocal regularization functional is assumed to be explicitly dependent on constant and pre-determined weights $\varphi$. The minimization problem here can be solved via a two-step alternating minimization scheme in an iterative projections onto convex sets (POCS) framework [23]. In each iteration, the first step projects the image estimate onto the data consistency term and the second step performs the minimization of the neighborhood penalty term on the projected data after re-estimating the weights from the current data estimate. The minimization of penalty term is equivalent to applying a non-local means (NLM) filter to the projected images [24]. The NLM filter is mathematically formulated as,

$$\hat{X}(p_x, p_y, p_t) = \frac{\sum_{(q_x, q_y, q_t) \in \mathcal{N}_P} \varphi(p, q) X(q_x, q_y, q_t)}{\sum_{(q_x, q_y, q_t) \in \mathcal{N}_P} \varphi(p, q)}, \quad (9)$$

and essentially calculates a weighted average of closest patches in a search neighborhood and updates every voxel accordingly. To reduce the computational burden of searching closest patches, we employed an optimized blockwise version of NLM proposed by Coupé *et al.* [25].

**Primal problem:** After solving each subproblem[1], we adopt a generalized forward-backward splitting (GFBS) framework [16] that jointly minimizes the sum of convex functions as given in our primal problem (2). GFBS is an operator-splitting algorithm and uses a forward-backward scheme [20]. Our proposed reconstruction algorithm is outlined in Algorithm 1. The algorithm mainly involves the computation of proximals on the gradient projections in every iteration and then weighted averaging of the two resulting proximal maps with weights denoted as $(w_1, w_2)$. We further accelerate the convergence of the algorithm with an additional acceleration step similar to the Fast Iterative Shrinkage-Thresholding Algorithm (FISTA) [26]. This step adaptively updates the value of step size parameter ($\alpha_k$) through iterations and make it sufficiently close to 1. The effect of adaptive $\alpha_k$ update on the convergence will be emprically demonstrated in Section IV-B1. The GFBS method has been shown to converge when $\gamma < 2/L$ if the convex function $f = \frac{1}{2}\|X - X_g\|_2^2$ has a Lipschitz continuous gradient with constant $L$. We refer the readers to GFBS paper [16] for more details concerning the proof of convergence.

---

[1]The details of algorithms solving each subproblem are given in supplementary material.

## III. DATA ANALYSIS

### A. Dynamic Susceptibility Contrast (DSC)

*1) Data:* Five glioma patients were imaged on a 3T MRI scanner using a 16-channel head neck coil. DSC-MRI image series were acquired using a 2D single-shot gradient-echo EPI sequence with parameters ($T_R = 1500$ ms, $T_E = 30$ ms, flip angle = 70°, voxel size = $1.8 \times 1.8 \times 4$ mm³, acquisition matrix = $128 \times 128$, 20 slices). A bolus of 15 ml Gd-DTPA (Magnevist, 0.5 mmol/ml) was injected 3 minutes after an initial first bolus of 7.5 ml with 4 ml/s injection rate. In total 60 frames were collected up to around 1.5 minutes.

*2) Analysis:* The signal time-intensity curves (TICs) of each voxel were directly used to estimate the CA concentration $C_{raw}$ from the change in the gradient echo transverse relaxation rate, $\triangle R_2^*$ [4],

$$C_{raw}(t) \propto \triangle R_2^* = - (1/T_E) \log [S(t)/S(0)], \quad (10)$$

where $S(t)$ is the post-contrast injection signal intensity, $S(0)$ is the pre-contrast signal intensity, and $T_E$ is echo time. Based on a well established tracer kinetic model in [3], the amount of contrast in the tissue is characterized by,

$$C_t(t) = \text{CBF} \cdot \int_0^t C_a(\tau) R(t - \tau) d\tau, \quad (11)$$

where $C_t(t)$ is the average CA concentration in a tissue voxel, CBF is the cerebral blood flow, $C_a(t)$ is the local CA concentration at the artery inlet, known as arterial input function (AIF), and $R(t)$ is the tissue residue function which measures the fraction of CA remaining in the given vascular network over time. The arterial input function $C_a(t)$ was determined over voxels in a small region extracted from branches of the middle cerebral artery. The noisy AIF signal was fit through a gamma-variate function to provide smooth concentration curves. Tissue residue functions $R(t)$ are obtained by deconvolving the tissue concentration time curves with the AIF using circulant truncated singular value decomposition [27]. The CBF was computed as the peak of the residue function, the CBV was determined as the area under the concentration time curves and the MTT was computed as the ratio of the CBV to CBF according to the central volume theorem [3].

### B. Dynamic Contrast-Enhanced (DCE)

*1) Data:* The data from five different glioblastoma patients were used for evaluation of the methods. For each patient, two data sets were sequentially acquired on a 3T MRI scanner: one for constructing $T_1$ maps and one for the DCE-MRI analysis. Data for constructing $T_1$ maps were acquired using a 3D fast gradient echo multiple flip angle approach with parameters ($T_R = 6$ ms, $T_E = 2.32$ ms, flip angles of $\{2°, 5°, 10°, 15°, 20°, 30°\}$, voxel size = $2 \times 2 \times 2$ mm³, acquisition matrix = $128 \times 128$, 20 slices). Dynamic DCE series were acquired with identical parameters but with a single flip angle of 10°. A bolus of 0.1 mmol/kg of GD-DTPA (gadopentetic acid) was injected after 52 s. Initially, a total of 250 frames were collected for up to 6 minutes. Each set of 5 frames were averaged before reconstruction, resulting 50 frames in total to constitute a dynamic sequence.

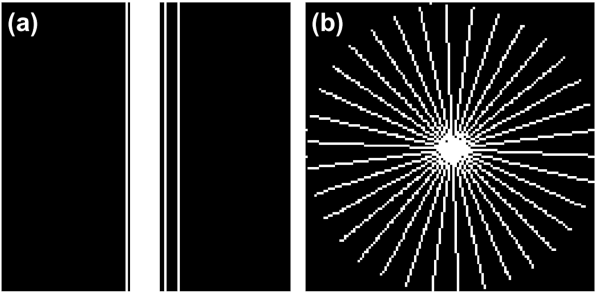

Fig. 2. Exemplary sampling patterns in (kx, ky) space corresponding to 8-fold acceleration: (a) Variable density Cartesian sampling (b) Radial Sampling.

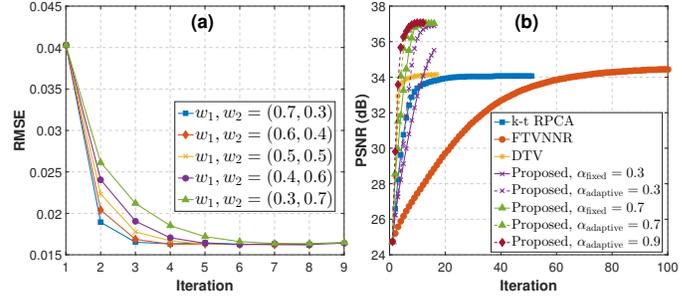

Fig. 3. (a) RMSE versus iteration number for different $(w_1, w_2)$ combinations obtained from proposed method, (b) PSNR versus iteration number for different reconstruction methods displaying the convergence of our algorithm depending on varying settings of step size ($\alpha$) parameter.

*2) Analysis:* Data collected at multiple flip angles for the $T_1$ map were fit using a non-linear least-squares fitting to the gradient echo signal intensity equation, given by [5]

$$S(t) = M \sin \alpha \frac{1 - \exp[-T_R/T_1(t)]}{1 - \cos \alpha \exp[-T_R/T_1(t)]}, \quad (12)$$

where $M$ is the proton density, $\alpha$ is the flip angle and $T_R$ is the repetition time. Here, we assume that $T_E \ll T_2^*$.

The increase in the relaxation rate can be further linearly related to the concentration of CA in the tissue, $C_t(t)$ [28],

$$1/T_1(t) = 1/T_1(0) + r_1 C_t(t), \quad (13)$$

where $T_1(0)$ is the tissue $T_1$ relaxation prior to the CA administration, $T_1(t)$ is the $T_1$ relaxation during and after injection, and $r_1$ is the CA relaxivity. We use the Patlak model [29] for tracer pharmacokinetic modeling and estimation of tissue perfusion parameters. This model describes a highly perfused two compartment tissue, ignoring backflux from the extracellular extravascular space (EES) into the blood plasma compartment. The CA concentration in the tissues is given by,

$$C_t(t) = v_p C_p(t) + K^{\text{trans}} \int_0^t C_p(\tau) d\tau, \quad (14)$$

where $K^{\text{trans}}$ is the volume transfer rate at which CA is delivered to the EES, $v_p$ is the plasma fraction, and $C_t(t)$ is the CA concentration in blood plasma. $C_p(t)$ denotes the AIF measured as in DSC analysis. The ROCKETSHIP toolbox [28] is used for $T_1$ map fitting and Patlak model implementation.

## IV. EXPERIMENTS AND RESULTS

### A. Experimental Setup

All perfusion datasets described in Section III were acquired as fully sampled data. The min-max normalized fully-sampled dynamic sequences were retrospectively undersampled by multiplying its corresponding k-space data with a binary undersampling mask and subsequently adding complex additive white Gaussian (AWG) noise, as formulated in (1). The power of AWG noise was fixed to $\sigma^2 = 10^{-10}$ in all experiments. Undersampling was simulated with a time-varying variable density Cartesian sampling and Radial sampling (see Fig. 2). These sampling strategies were commonly used for dynamic MR applications [12], [9], [10]. To provide better evaluation of different methods, we considered increasing acceleration factors of R = {2x, 4x, 8x, 12x, 16x} in the experiments. The reference image used in the $\mathcal{R}_L$-subproblem was initially taken as the direct inverse FFT (zero-filled) reconstruction of the first frame with 2-fold subsampling. Later on, it was updated as the average of all frames in the reconstructed sequence through iterations. Perfusion parameter estimation was employed after image reconstruction as a separate step.

We compare our method with three state-of-the-art dynamic reconstruction techniques: (k,t)-space via low-rank plus sparse prior (k-t RPCA) [12], dynamic total variation [9], fast total variation and nuclear norm regularization (FTVNNR) [6]. To ensure fair comparison, as presented similarly in [10], we empirically fine-tuned the optimal regularization parameters for all three methods and individually for each dataset considering the suggested parameter space in the respective papers. The regularization parameters of our algorithm were set as $\lambda_1 = 0.001$ and $\lambda_2 = 0.25$. We also fixed $N_w = 7 \times 7 \times 7$ and $N_p = 5 \times 5 \times 5$ in all experiments. We considered using small cubic patches for $N_w$ and $N_p$ because larger patches drastically increase the computation time despite not improving the results substantially. The choice of proximal weights $w_1, w_2$ and step size $\alpha$ parameters of Algorithm 1 will be explained in Section IV-B1.

The quality of reconstructions was quantitatively measured with the Root-Mean-Square Error (RMSE) and Peak Signal-to-Noise Ratio (PSNR) metrics. RMSE for a 3D reconstructed sequence involving in total $N$ voxels is calculated as $\text{RMSE}(X_r) = \sqrt{\|X_f - X_r\|_2^2/N}$ and PSNR is computed as $\text{PSNR}(X_r) = 20 \log [1/\text{RMSE}(X_r)]$, where $X_f$ denotes the fully sampled sequence. To ensure convergence, all reconstruction methods were stopped when a maximum number of iterations (varying depending on the method) was reached, or when $(\|X_r^{k+1} - X_r^k\|_2^2)/\|X_r^k\|_2^2 \leq 10^{-6}$, where $k$ is the iteration number. For the purposes of evaluation, we treated the fully sampled data as ground truth. As a commonly used metric in quantitative imaging [27], [7] we adopted the Lin's Concordance Correlation Coefficients (CCCs) to quantitatively assess the agreement of estimated perfusion parameters with reference values obtained from ground truth.

### B. Results

*1) Convergence of Proposed Algorithm and Effects of Parameters:* In this experiment, we investigate the effect of proximal weights $(w_1, w_2)$ and step size ($\alpha$) parameters on the convergence speed of our proposed algorithm. Fig. 3 shows the resulting RMSE and PSNR values versus iterations depending on various combinations of $(w_1, w_2)$ pairs and varying settings of $\alpha$. Fig. 3(a) reveals that the convergence rate of our proposed algorithm increases when the weight of the first proximal has a higher value. The highest convergence rate is achieved by the weight pair of $(0.7, 0.3)$. This implies that $\mathcal{R}_\text{L}$-proximal yields better intermediate reconstruction accuracy than the one obtained by $\mathcal{R}_\text{NL}$-proximal. Therefore, a higher weight given to $\mathcal{R}_\text{L}$-proximal will yield lower RMSE in a less number of iterations, indicating higher convergence rate. Fig. 3(a) also shows that all combinations of weights ultimately reach similar RMSE values, meaning that the final reconstructions are not affected by different proximal weight-ing. This demonstrates the stability and robustness of the GFBS based splitting algorithm against proximal weighting.

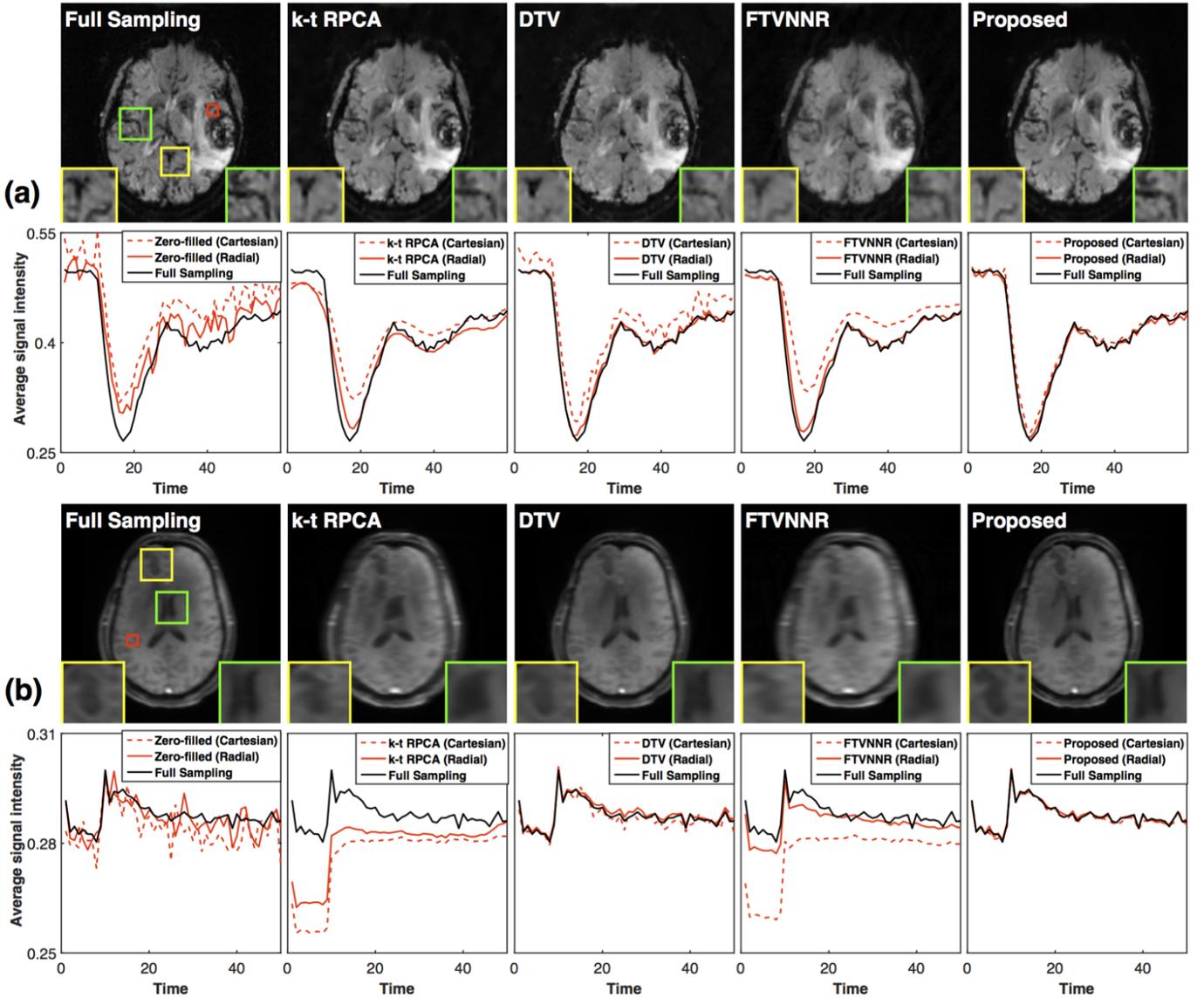

Fig. 4. Qualitative reconstruction results of a single frame and perfusion time intensity curves (TICs) obtained by each reconstruction algorithm. (a) DSC subject data. (b) DCE subject data. Single frame reconstructions are obtained using radial sampling with 12-fold acceleration for DSC data and 8-fold acceleration for DCE data. Close-up views of two regions of interest (yellow and green square) are also provided in every single frame reconstruction images. The TICs display the signal intensity over time averaged over the voxels inside the red square (corresponding to an arterial region) as shown in Full Sampling images, obtained from using Cartesian and Radial sampling schemes. Our proposed model results in high quality image frames as observed in close-up views. The reconstructed TICs also show strong alignment with the ones obtained from fully sampled data when undersampled by two different sampling schemes.

The impact of various step size ($\alpha$) parameter settings on reconstruction performance is displayed in Fig. 3(b). In this experiment, we set $w_1, w_2 = (0.5, 0.5)$ for all cases of $\alpha$. $\alpha_\text{fixed}$ denotes a fixed $\alpha$ value while $\alpha_\text{adaptive}$ denotes adaptive $\alpha$ updated through iterations as given in Algorithm 1. The results in Fig. 3(b) reveal that our proposed adaptive setting of $\alpha$ increases the convergence speed of the algorithm. This is clearly visible when $\alpha_\text{fixed} = \alpha_\text{adaptive} = 0.3$. After 6th iteration, our proposed method reaches the highest PSNR value (with $\alpha_\text{adaptive} = 0.3$) when compared to other methods. Fig. 3(b) also demonstrates that the acceleration gain of convergence decreases when initial $\alpha_0$ has a higher value, for instance

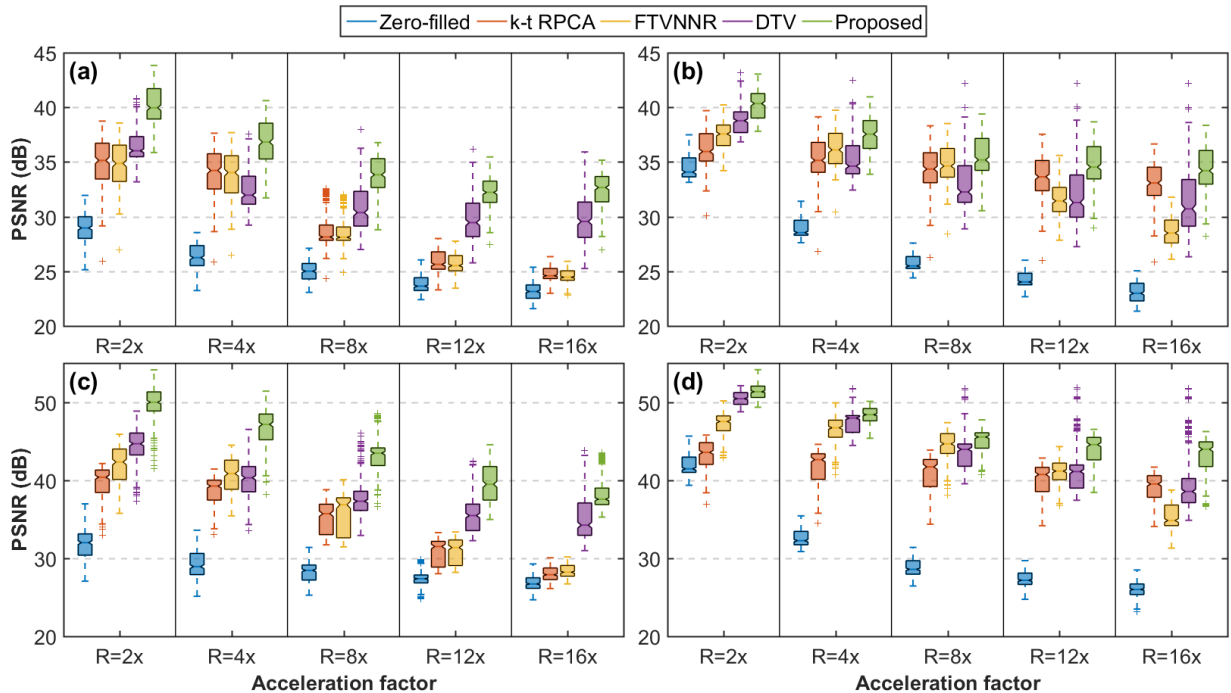

Fig. 5. Boxplots displaying the PSNR values of reconstruction methods with respect to increasing acceleration factors R. Reconstruction results obtained from all 5 DSC sequences with (a) Cartesian, and (b) Radial sampling; from all 5 DCE sequences with (c) Cartesian, and (d) Radial sampling. Each DSC sequence involves 60 frames and each DCE sequence involves 50 frames. PSNR values were calculated for every single frame individually. The proposed method performs the best average PSNR in all acceleration rates for both with DSC and DCE datasets.

$\alpha_{\text{fixed}} = \alpha_{\text{adaptive}} = 0.7$. The PSNR results obtained with $\alpha_{\text{adaptive}} = 0.9$ suggests that increasing step-size parameter of GFBS algorithm enables faster convergence. To this end, we set the initial $\alpha_0 = 0.9$ and use the adaptive setting scheme for $\alpha$ in the remaining experiments of the paper. This allows us to reach the highest reconstruction accuracy in the least number of iterations, thereby significantly reducing processing time.

*2) Reconstruction Performance:* This section presents the reconstruction results of all competing methods using variable density Cartesian and radial sampling schemes. Fig. 4 demonstrates a single reconstructed frame of one of the DSC and DCE brain perfusion datasets and estimated perfusion TICs averaged over voxels inside an arterial region. The acceleration factors here are 12 and 8 for DSC and DCE data, respectively. The results in Fig. 4 show that our proposed reconstruction algorithm can achieve the best spatial reconstruction and highly accurate estimation of TICs when compared to the other three methods. Considering the spatial results, when looking at details in close-up views of Fig. 4(a), DSC frame reconstructions obtained by k-t RPCA and our proposed method produce the best results compared to DTV and FTVNNR. DTV reconstructions are mostly lacking finer details whereas FTVNNR yields more blurry spatial regions. The TIC reconstructions in Fig. 4(a) indicate that radial sampling yields more accurate matching of full sampling TICs compared to variable density Cartesian sampling. Among all TIC results, DTV and our proposed method reconstruct perfusion signal patterns that are in good agreement with the pattern of the fully sampled data (see Fig. 4(a) bottom third and fifth column) when radial sampling is used for undersampling. DTV produces signifi-cantly worse TICs with underestimated perfusion peaks while our method still yields very accurate matching of TICs with Cartesian sampling. As we also demonstrated in our previous work [17], k-t RPCA and FTVNNR estimate oversmooth TICs and reconstruct underestimated perfusion peaks with Cartesian sampling (see Fig. 4(a) bottom second and fourth column). However, radial sampling helps these two methods to improve their accuracy in TIC estimation, especially FTVNNR yields highly accurate matching of TICs with fully sampled data. These results evidence that radial sampling should be preferred over Cartesian sampling for the quantitative perfusion MRI in which the fidelity of TICs plays a vital role.

Considering the spatial results of DCE data in Fig. 4(b), DTV and our proposed method provide the best reconstructions. When looking at details in close-up views, it is visible that our method reconstructs finer details compared to DTV. FTVNNR again shows more blurry regions and thus lacking details. Unlike the results in DSC data, k-t RPCA yields spatial reconstructions with missing details and blurred edges in DCE data. In terms of TIC reconstruction, Fig. 4-(b) demonstrates that our method can achieve the most accurate TICs both with Cartesian and radial sampling, where DTV achieves the second best. However, when the observed signal dynamics are lower compared to DSC, both k-t RPCA and FTVNNR fail in the estimation of perfusion TICs in DCE. The peaks of the perfusion signal are underestimated and there is a large offset between the estimation and real value. This result reveals that k-t RPCA and FTVNNR are not robust against small temporal variations since these methods do not explicitly exploit sparsity in the temporal domain while our method exploits the temporal

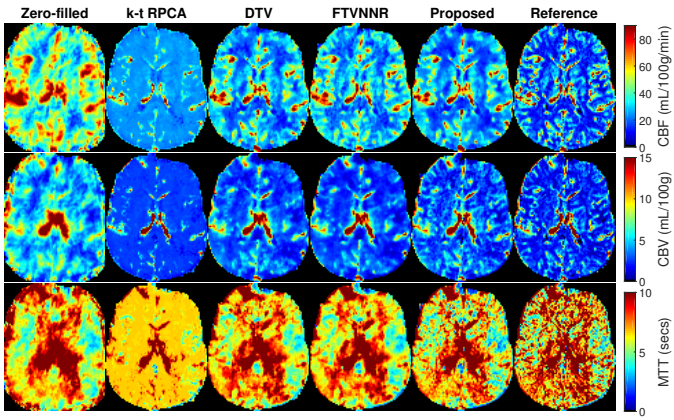

Fig. 6. Hemodynamic parameter maps (CBF, CBV, MTT) of a DSC subject for different methods with an 8-fold acceleration. This subject has a low-grade glioma which cannot be easily recognized.

variations in both voxel and patch-wise levels.

Fig. 5 presents the quantitative results (in terms of PSNR) of all reconstruction methods depending on increasing acceleration factors for both DSC and DCE sequences. As expected, PSNR values decrease with increasing acceleration due to the increase in missing k-space samples. Moreover, all reconstruction methods benefit from radial sampling because it can be seen that zero-filled reconstruction already gives an improved reconstruction compared to the one obtained by Cartesian sampling. The reason is that the resulting artifacts produced by radial sampling resemble noise compared to Cartesian undersampling [12]. Fig. 5 clearly shows that our proposed method performs the best PSNR in all acceleration rates for both with DSC and DCE datasets. It is also observed that the performance of k-t RPCA and FTVNNR is improved with radial undersampling while on average DTV performs worse with radial sampling especially in higher accelerations.

*3) Perfusion Parameter Estimation:* For the estimation of perfusion parameters, we only used the reconstruction results obtained by radial sampling due to its efficiency in both spatial and temporal reconstruction, as demonstrated in the previous section. Figs. 6-7 display qualitative results of estimated hemodynamic parameter maps from two DSC subjects. From Fig. 6, it can be seen that the proposed method results in perfusion maps where most of the tissue structures are preserved and appear sharper compared to FTVNNR and DTV with 8-fold acceleration. Compared to reference maps, CBF and CBV values are overestimated in some regions of white matter (WM) and gray matter (GM) while perfusion values in blood vessels (appear as small red areas) stay mostly accurate. The k-t RPCA method produces highly inaccurate (oversmooth) perfusion maps as expected by the mismatch of TICs shown in Fig. 4(a). Fig. 7 demonstrates that estimated hemodynamic parameter maps generated by our proposed method appear highly accurate up to 8-fold acceleration, however the maps start to deteriorate and involve oversmooth regions at higher acceleration rates. CBF assessment in tumor areas (at tumor boundary and core) also appear quite consistent with reference maps up to 12-fold acceleration.

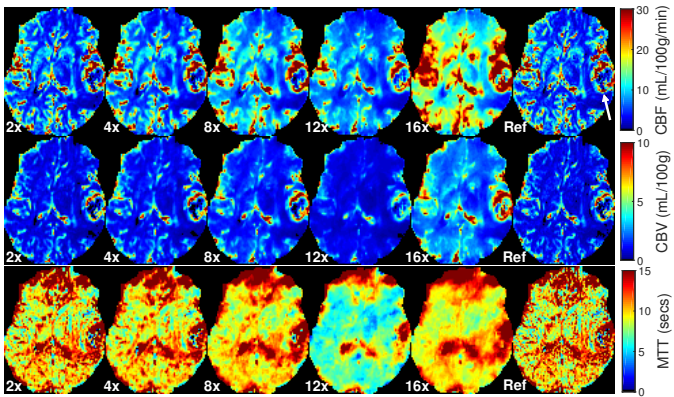

Fig. 7. Hemodynamic parameter maps (CBF, CBV, MTT) of a DSC subject resulting from our proposed method with respect to different acceleration factors and Reference (Ref) maps for comparison. White arrow in Reference CBF map indicates the tumor region.

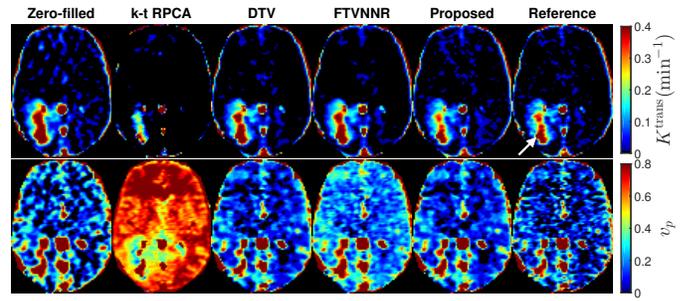

Fig. 8. Pharmacokinetic parameter maps ($K^{trans}$, $v_p$) of a DCE subject for different methods with a 12-fold acceleration. The tumor area is marked by a white arrow in the Reference $K^{trans}$ map.

Fig. 8 shows estimated pharmacokinetic parameter maps of a DCE subject with a 12-fold acceleration. The estimated maps indicate that significantly higher permeability ($K^{trans}$) can be observed in tumor tissues (white arrow). The proposed method produces parameter maps which show a strong match with reference maps at the tumor region and exhibit relatively oversmooth regions in WM and GM due to a high rate of acceleration (12-fold). The k-t RPCA especially produces highly overestimated plasma fraction $v_p$ values and again reveals the fact that it is inadequate for PWI reconstruction. Due to the page limitation, we provide the qualitative results of an another DCE subject in supplementary material.

Fig. 9 presents Bland-Altman plots of the CBF and $K^{trans}$ values obtained by our method with respect to increasing acceleration factors. The plots indicate that the fidelity of estimation is very high at lower acceleration rates where we observe very small differences between the estimation and ground truth. With increasing acceleration, the bias and variance of the differences generally become larger and subsequently CCC values diminish. This quantitative assessment coincides with the qualitative results shown in Fig. 7.

Finally, Table I reports the average CCCs of DSC and DCE perfusion parameter maps obtained from all methods with varying acceleration rates. Our reconstruction method yields the best CCCs for all parameters and accelerations. The k-t RPCA generally performs even worse than zero-filled reconstruction since it leads to oversmoothing of the temporal perfusion signal. The DTV and FTVNNR show relatively similar performance with our method at lower accelerations

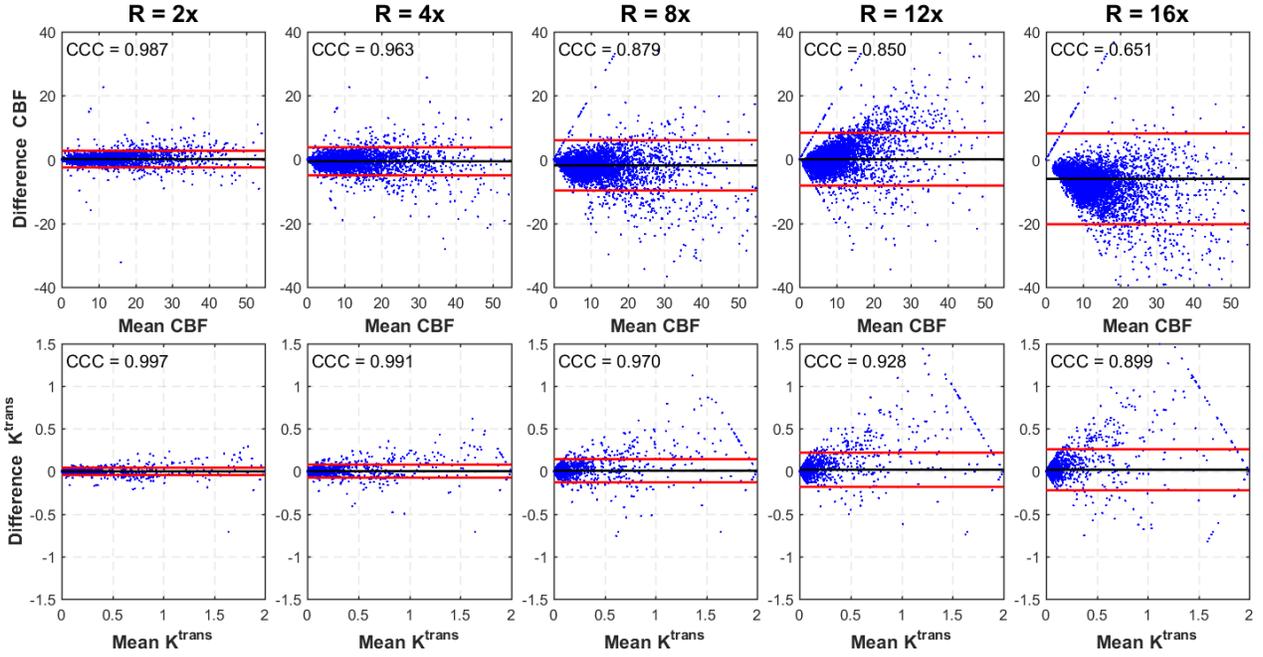

Fig. 9. Bland-Altman plots and 95% confidence intervals within two red lines for CBF (top) and $K^{trans}$ (bottom) perfusion parameters of one DSC and DCE subject data depending on different acceleration factors, resulting from our proposed reconstruction method. Corrresponding CCCs are also provided at the top-left corner of each plot. For these subjects, CBF values can range from 0 to 71 mL/100g/min and $K^{trans}$ values can range from 0 to 2 min$^{-1}$. As can be clearly seen in the plots, CCCs decrease with increasing acceleration, which coincides with the changes of bias and variance.

TABLE I
AVERAGE CCC VALUES OF THE DSC PERFUSION MAPS (CBF, CBV, MTT) AND DCE PERFUSION MAPS ($K^{trans}$, $v_p$) ESTIMATION OBTAINED FROM
ALL THE SUBJECTS. CCCS TAKE VALUES IN $[0, 1]$ INTERVAL. THE BEST PERFORMANCE IS HIGHLIGHTED IN BOLD FONT.

| | | Acceleration Rate | | | | |
|---|---|---|---|---|---|---|
| | Method | R=2x | R=4x | R=8x | R=12x | R=16x |
| DSC Parameters | Zero-filled | 0.924, 0.888, 0.854 | 0.718, 0.684, 0.630 | 0.485, 0.479, 0.432 | 0.345, 0.362, 0.320 | 0.221, 0.205, 0.155 |
| | k-t RPCA | 0.839, 0.786, 0.767 | 0.889, 0.897, 0.796 | 0.812, 0.823, 0.728 | 0.733, 0.719, 0.654 | 0.634, 0.625, 0.644 |
| | DTV | 0.962, 0.957, 0.939 | 0.930, 0.930, 0.889 | 0.807, 0.804, 0.813 | 0.713, 0.691, 0.751 | 0.614, 0.628, 0.734 |
| | FTVNNR | 0.966, 0.941, 0.870 | 0.946, 0.937, 0.841 | 0.804, 0.784, 0.733 | 0.695, 0.689, 0.614 | 0.603, 0.600, 0.429 |
| | Proposed | **0.968, 0.962, 0.947** | **0.958, 0.948, 0.897** | **0.887, 0.862, 0.821** | **0.741, 0.751, 0.759** | **0.637, 0.663, 0.739** |
| DCE Parameters | Zero-filled | 0.987, 0.973 | 0.927, 0.857 | 0.877, 0.727 | 0.763, 0.677 | 0.741, 0.663 |
| | k-t RPCA | 0.790, 0.681 | 0.634, 0.370 | 0.596, 0.352 | 0.531, 0.312 | 0.498, 0.275 |
| | DTV | 0.990, 0.985 | 0.943, 0.941 | 0.847, 0.813 | 0.789, 0.736 | 0.764, 0.670 |
| | FTVNNR | 0.984, 0.903 | 0.952, 0.839 | 0.879, 0.694 | 0.848, 0.589 | 0.791, 0.546 |
| | Proposed | **0.994, 0.991** | **0.973, 0.967** | **0.919, 0.845** | **0.889, 0.768** | **0.848, 0.686** |

(up to 4-fold) and worse performance at higher accelerations.

*4) Computation Time:* The most computationally expensive step of our algorithm is solving each proximal map. We process the computation of proximal maps in parallel since proximal-splitting can allow it due to the independence between the inputs of proximity operators. Other steps of the algorithm involve adding and multiplying vectors or scalars, and are thus very cheap in terms of computational complexity.

All methods were ran using Matlab R2015b on a desktop computer with Intel Xeon CPU E3-1226 v3 Processor at 3.3 GHz and 32 GiB of memory. Table II provides the processing time (in seconds) of all methods on two types of datasets. Among all methods, our method requires the highest processing time. However, within a similar computation time (corresponding to $\simeq$ 3-4 iterations) as competing methods, our method can usually reach the best reconstruction accuracy.

Considering the longer processing times ($\simeq$ 15 minutes for DCE data analysis) for voxel-wise fitting of perfusion parameters, we believe that the slightly longer reconstruction time of our method can be negligible.

## V. DISCUSSION AND CONCLUSION

In this paper, a new reconstruction model exploiting spatiotemporal variations jointly at multiple levels was proposed for the acceleration of PWI acquisitions. The proposed method was compared with existing state-of-the-art reconstruction methods and evaluated on clinical DSC and DCE-MRI patient datasets. Extensive experiments validated the effectiveness of our method in terms of improved spatial reconstructions, highly accurate matching of perfusion temporal signals, and more precise estimation of clinically relevant perfusion parameters. Experiments based on retrospective undersampling revealed that our reconstruction model can potentially enable up to 8-fold acceleration on clinically feasible perfusion datasets. We also demonstrated that our method is very robust against incoherent artifacts caused by varying sampling patterns. Our recent work [30] showed that the proposed model can also achieve similar performances with Poisson-disc sampling.

TABLE II
COMPUTATION TIME OF DIFFERENT RECONSTRUCTION METHODS.

| Time (secs) | k-t RPCA | DTV | FTVNNR | Proposed |
|---|---|---|---|---|
| DSC dataset | 187.2 | 72.3 | 142.4 | 324.1 |
| DCE dataset | 165.6 | 61.8 | 106.3 | 268.7 |

The maximum acceleration achieved with our method can be further increased with the use of a high-spatial-resolution data as applied in [7], [15] for DCE study. However, we remark that high spatial resolution is not so clinically realistic for PWI because high temporal resolution is vital to capture entire contrast dynamics for precise blood flow quantification.

As mentioned already, this paper considers the reconstruction on $2D + t$ data, i.e., on a single slice followed over time. However, our approach can be easily extended to $3D + t$ data. One should take into account time complexity because especially applying NLM filter on $4D$ patches can significantly increase the computation time due to exhaustive search of similar patches in larger windows. Hence, an optimized GPU implementation is necessary to perform it efficiently in $4D$.

We would like to emphasize that our regularization approach can also be adopted to many inverse problems in medical imaging with only a few modifications. Possible applications beyond perfusion imaging might be MR Spectroscopy [31], low-dose CT denoising [32], and MR super-resolution [33]. Moreover, our algorithm can provide an efficient way of solving such regularized inverse problems.

One of the limitation of our method is that it usually produces oversmooth reconstructions and therefore overestimated perfusion parameters when the acceleration factor is relatively higher ($\geq$12-fold). This observation is mostly valid for parenchyma voxels (i.e., WM and GM) where the signal drop or enhancement is very low compared to blood vessels. Iterative reconstruction algorithms usually tend to produce repeating structures and smooth the available information throughout neighboring image regions when there is a large amount of missing data due to the high undersampling. In order to obtain more accurate parameter estimates in highly undersampled data, deep neural networks [34] can be exploited to learn deeper spatio-temporal representations and similarities within the MR image series than conventional NLM can learn with a simple $k$-nearest neighbors approach. One idea might be to learn the noise-free temporal signal for every voxel within the spatially correlated image regions given the corresponding noisy signal using deep learning methods. Future lines of research will attempt to explore this idea.

## VI. SUPPLEMENTARY MATERIAL

In this supplementary, we provide a few extra materials that have been referenced in the original manuscript.

### A. A Diagram on parameter estimation in DSC-MRI

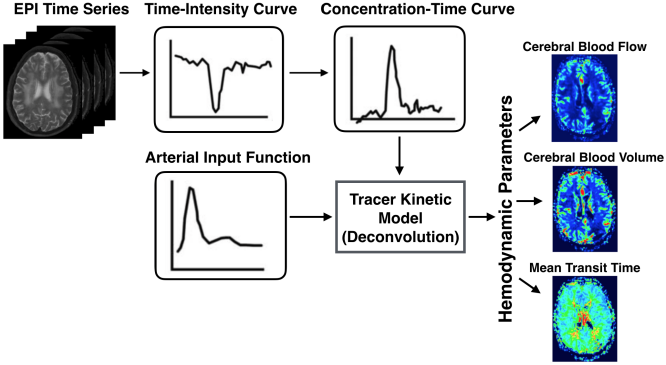

Fig. 10. A diagram illustrating the steps of hemodynamic parameter estimation in dynamic susceptibility contrast (DSC) $T_2^*$-weighted perfusion MRI. First, time-intensity curves (TICs) for each voxel are converted to tracer tissue concentration-time courses based on a linear relationship between the signal drop in $T_2^*$ and tissue concentration. Then, a tracer kinetic model based on the deconvolution of arterial input function and tissue concentrations is applied to determine perfusion parameters [4]. As the voxel-wise TICs are directly correlated to the amount of contrast medium in tissue, precise estimation of TICs is very crucial to accurately quantify the hemodynamic parameters. Especially under and over estimation of the peak of the TICs entirely reflect the resulting parameters obtained by a deconvolution process.

### B. Algorithm for solving $\mathcal{R}_L$-subproblem

As stated in the manuscript, this subproblem can be solved efficiently using an accelerated iteratively reweighted least squares algorithm proposed for structured sparsity reconstruction [21]. The total-variation (TV) term in our problem (7) can be modeled as the group sparsity,

$$\|x\|_{TV} = \|x\|_{2,1} = \sum \|x_{g_i}\|_2, \ i = 1, 2, ..., m \quad (15)$$

where $x_{g_i}$ denotes the component in the $i$-th group and $m$ is the total number of groups. For instance, TV component in each dimension (e.g. $\nabla_x x$) constitutes a group. Following the notations given in (II-B), the problem in (7) is solved by iteratively updating the weight matrix $W$ and the solution $d_t$ for each frame individually [9]. $W$ is a diagonal matrix with the $i$-th diagonal entry given as

$$W_i^k = 1 / \sqrt{\left(\nabla_x d_{t_i}^k\right)^2 + \left(\nabla_y d_{t_i}^k\right)^2}, \ i = 1, 2, ..., N_x N_y \quad (16)$$

where $k$ is the iteration number. $d_t^{k+1}$ is updated by solving the following linear equation:

$$\left(\mathcal{F}_t^H \mathcal{F}_t + \lambda Q_1^H W^k Q_1 + \lambda Q_2^H W^k Q_2\right) d_t = \mathcal{F}_t^H b_t, \quad (17)$$

where $b_t = y_t - \mathcal{F}_t \bar{x}$. The close form solution of (17) is derived as $d_t^k = S^{-1} \mathcal{F}_t^H b_t$, where $S = \mathcal{F}_t^H \mathcal{F}_t + \lambda Q_1^H W^k Q_1 + \lambda Q_2^H W^k Q_2$ is the system matrix. However, the direct inversion of $S$ is not computationally feasible. In [21], it is proposed to use a preconditioner $P$ which is close to $S$ and the inverse of $S$ can be computed in a more computationally efficient way. The overall problem in (17) can be solved by a preconditioned conjugate gradient (PCG) method.

---

**Algorithm 2:** $\mathcal{R}_L$-subproblem

**Input:** $\mathcal{F}_u$, $Y$, Initial estimate $X^0$, $\lambda = 2\lambda_1$
**for** $t \in \mathbb{T} = \{1, 2, ..., T\}$ **do**
  **Initialize:** $d_t^0 = x_t^0 - \bar{x}$, $b_t = y_t - \mathcal{F}_t \bar{x}$, $k = 0$
  **while** *stopping criteria not met* **do**
    Obtain $W^k$ via (16) ;
    $S = \mathcal{F}_t^H \mathcal{F}_t + \lambda Q_1^H W^k Q_1 + \lambda Q_2^H W^k Q_2$ ;
    $P = sI + \lambda Q_1^H W^k Q_1 + \lambda Q_2^H W^k Q_2 = LU, P^{-1} = U^{-1} L^{-1}$ ;
    **while** *PCG stopping criteria not met* **do**
      Update $d_t^{k+1}$ by PCG for $Sd_t = \mathcal{F}_u^H b_t$ with $P \approx LU$ ;
    **end**
    $k \leftarrow k + 1$ ;
  **end**
  $\hat{x}_t = d_t + \bar{x}$ ;
**end**
**Output:** $\hat{X}_{\mathcal{R}_L} = [\hat{x}_1, \hat{x}_2, \cdots, \hat{x}_t]$

---

The preconditioner $P$ in our problem can be designed based on the following observation [9]: The symmetric matrix $A_t^H A_t$ is diagonal and therefore $\mathcal{F}_t^H \mathcal{F}_t = F^H A_t^H A_t F$ is diagonally dominant. Due to the properties of the Fourier transform, all the diagonal elements of $\mathcal{F}_t^H \mathcal{F}_t$ is equal to the mean of diagonal elements of $A_t^H A_t$, which is the undersampling factor denoted as $s$. A good approximation for $\mathcal{F}_t^H \mathcal{F}_t$ can be made with $sI$, where $I$ is the identity matrix. The new preconditioner is finally defined as $P = sI + \lambda Q_1^H W^k Q_1 + \lambda Q_2^H W^k Q_2$. The new procoditioner $P$ is a symmetric penta-diagonal matrix which does not have a closed form inverse. However, $P$ is usually diaogonally dominant because the regularization parameter $\lambda$ is often very small in our specific problem. Therefore, an incomplete LU decomposition can be applied to such matrix with $P \approx LU$, where $L$ and $U$ are a lower triangle matrix and an upper triangle matrix, respectively.

Provided all the details, the steps of algorithm solving $\mathcal{R}_L$-subproblem is outlined in Algorithm 2.

### C. Algorithm for solving $\mathcal{R}_{NL}$-subproblem

The subproblem is solved via a simple two-step alternating minimization scheme. In every main iteration, the reconstructed image data estimate is first projected onto the data fidelity term and then NLM filtering is applied to the projected data in entire spatio-temporal (3D) space. The steps of the algorithm solving $\mathcal{R}_{NL}$-subproblem is provided below in Algorithm 3.

In this algorithm, $NLM$ operation performs the NLM filtering on the projected data given the filter parameters. Basically, in each spatio-neighborhood window $\mathcal{N}_\mathbf{p}$, first the

---

**Algorithm 3:** $\mathcal{R}_{NL}$-subproblem

**Input:** $\mathcal{F}_u$, $Y$, Initial estimate $X^0$, $\alpha = 2\lambda_2$, $N_w$, $N_p$, $\sigma$
**Initialize:** $X_{est}^0 = X^0$, $h = 0.2\sigma$, $k = 0$
**while** *stopping criteria not met* **do**
  $X_{proj} = X_{est}^k + \mathcal{F}_u^H \left(Y - \mathcal{F}_u X_{est}^k\right)$ ;
  $X_{nlm} = NLM(X_{proj}, N_w, N_p, h)$ ;
  $X_{est}^{k+1} = X_{est}^k + \alpha(X_{nlm} - X_{est}^k)$ ;
  $k \leftarrow k + 1$ ;
**end**
**Output:** $\hat{X}_{\mathcal{R}_{NL}} = X_{est}$

weights between the centers of the patches are calculated via (3) and then these weights are directly used to update each voxel value via (9) taking into account the other voxels in $\mathcal{N}_{\mathbf{p}}$ surrounding the voxel $\mathbf{p}$. As the inter-pixel weights are re-estimated in every iteration, the estimated weights become more reliable when the quality of the reconstructed data $X_{\text{est}}$ are improved through iterations.

We note that the classical NLM filter [24] was normally defined with a Gaussian-weighted Euclidean distance, $\|\cdot\|_{2,a}^2$, where $a$ is the standard deviation of a Gaussian kernel. This kernel basically gives decaying weights to voxel differences away from the center of the patches. However, in this work, we used the classical Euclidean distance $\|\cdot\|_2^2$ to simplify the complexity of the problem and reduce the computational time as proposed in [25]. We anticipate that a Gaussian-weighted distance measure may slightly improve the quality of our reconstructions because assigning uniform weights to all the voxels inside a patch may lead to stronger smoothing in image regions involving especially large texture and more fine details.

### D. Additional Qualitative Parameter Maps of a DCE Subject

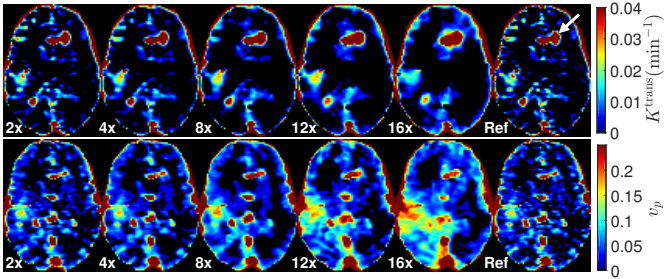

Fig. 11. Pharmacokinetic parameter maps ($K^{\text{trans}}, v_p$) of a DCE subject obtained by our proposed method with respect to increasing acceleration factors and Reference (Ref) maps for comparison. White arrow in Reference $K^{\text{trans}}$ map marks the tumor region.

Fig. 11 shows the estimated pharmacokinetic parameter maps of our proposed reconstruction model with respect to increasing acceleration factors. The estimated maps look very similar up to 4-fold acceleration when compared to reference maps. However, starting from 8-fold acceleration, our method yields oversmooth image regions in different brain areas. Especially in higher acceleration factors ($\geq$12-fold), $K^{\text{trans}}$ values are largely underestimated in WM and GM, and the maps contain expanded tumor regions (oversmoothed). On the other hand, $v_p$ values are mostly overestimated in WM and GM due to oversmoothing, and we clearly observe expanded regions especially around a few blood vessels which are highly perfused and affect the neighboring brain areas as well.